\documentclass[pdflatex,sn-nature]{sn-jnl}

\usepackage{graphicx}%
\usepackage{multirow}%
\usepackage{amsmath,amssymb,amsfonts}%
\usepackage{amsthm}%
\usepackage{mathrsfs}%
\usepackage[title]{appendix}%
\usepackage{xcolor}%
\usepackage{textcomp}%
\usepackage{manyfoot}%
\usepackage{booktabs}%
\usepackage{algorithm}%
\usepackage{algorithmicx}%
\usepackage{algpseudocode}%
\usepackage{listings}%
\usepackage{todonotes}
\usepackage{comment}
\usepackage{amsmath,amssymb,amsthm}
\usepackage{bm}
\usepackage{stmaryrd}

\usepackage{listings}
\usepackage{xcolor}
\usepackage{float}
\usepackage{caption}

\newtheorem*{definition*}{Definition} 
\lstdefinelanguage{SPARQL}{
  keywords={SELECT, WHERE, VALUES, FILTER, NOT, EXISTS, DISTINCT, PREFIX},
  keywordstyle=\bfseries\color{blue},
  sensitive=true,
  comment=[l]{\#},
  commentstyle=\color{green!50!black},
  string=[b]{"},
  stringstyle=\color{red},
  morestring=[b]{'},
  morecomment=[l]{//},
  morecomment=[s]{/*}{*/},
}

\lstdefinestyle{sparqlstyle}{
  language=SPARQL,
  basicstyle=\ttfamily\small,
  numbers=left,
  numberstyle=\tiny,
  stepnumber=1,
  numbersep=5pt,
  backgroundcolor=\color{gray!10},
  showspaces=false,
  showstringspaces=false,
  showtabs=false,
  frame=single,
  tabsize=2,
  captionpos=b,
  breaklines=true,
  breakatwhitespace=false,
  escapeinside={(*@}{@*)},
  framesep=5pt,
  xleftmargin=15pt,
  xrightmargin=15pt,
}
\graphicspath{{Fig/}}

\theoremstyle{thmstyleone}%
\theoremstyle{thmstyletwo}%

\theoremstyle{thmstylethree}%
\usepackage{todonotes}
\usepackage{comment}
\usepackage{amsmath,amssymb,amsthm}
\usepackage{bm}
\usepackage{stmaryrd}

\usepackage{listings}
\usepackage{xcolor}
\usepackage{float}
\usepackage{caption}

\begin{document}

\title[]{Causal knowledge graph analysis identifies adverse drug effects}


\author[1]{\fnm{Sumyyah} \sur{Toonsi}}\email{sumyyah.toonsi@kaust.edu.sa}

\author[2]{\fnm{Paul} \sur{N. Schofield}}\email{pns12@cam.ac.uk}

\author*[1,3,4]{\fnm{Robert} \sur{Hoehndorf}}\email{robert.hoehndorf@kaust.edu.sa}

\affil*[1]{\orgdiv{Computer, Electrical and Mathematical Sciences \&
    Engineering}, \orgname{King Abdullah University of Science and
    Technology}, \orgaddress{\street{4700 KAUST}, \postcode{23955},
    \state{Thuwal}, \country{Saudi Arabia}}}
\affil[2]{\orgdiv{Department of Physiology, Development \&
    Neuroscience}, \orgname{University of Cambridge},
  \orgaddress{\street{Downing Street}, \postcode{CB2 3EG},
    \country{United Kingdom}}}

\affil[3]{\orgdiv{SDAIA--KAUST Center of Excellence in Data Science
    and Artificial Intelligence}, \orgname{King Abdullah University of
    Science and Technology}, \orgaddress{\street{4700 King Abdullah
      University of Science and Technology}, \state{Thuwal},
    \country{Saudi Arabia}}}

\affil[4]{\orgdiv{KAUST Center of Excellence for Smart Health
    (KCSH)}, \orgname{King Abdullah University of Science and
    Technology}, \orgaddress{\street{4700 King Abdullah University of
      Science and Technology}, \state{Thuwal}, \country{Saudi
      Arabia}}}


\abstract{Knowledge graphs and structural causal models have each
  proven valuable for organizing biomedical knowledge and estimating
  causal effects, but remain largely disconnected: knowledge graphs
  encode qualitative relationships focusing on facts and deductive
  reasoning without formal probabilistic semantics, while causal
  models lack integration with background knowledge in knowledge
  graphs and have no access to the deductive reasoning capabilities
  that knowledge graphs provide.
  To bridge this gap, we introduce a novel formulation of Causal
  Knowledge Graphs (CKGs) which extend knowledge graphs with formal
  causal semantics, preserving their deductive capabilities while
  enabling principled causal inference.
  CKGs support deconfounding via explicitly marked causal edges and
  facilitate hypothesis formulation aligned with both encoded and
  entailed background knowledge. 
  We constructed a Drug--Disease CKG (DD-CKG) integrating disease
  progression pathways, drug indications, side-effects, and
  hierarchical disease classification to enable automated large-scale
  mediation analysis.  Applied to UK Biobank and MIMIC-IV cohorts, we
  tested whether drugs mediate effects between indications and
  downstream disease progression, adjusting for confounders inferred
  from the DD-CKG. Our approach successfully reproduced known adverse
  drug reactions with high precision while identifying previously
  undocumented significant candidate adverse effects.  Further
  validation through side effect similarity analysis demonstrated that
  combining our predicted drug effects with established databases
  significantly improves the prediction of shared drug indications,
  supporting the clinical relevance of our novel findings. These
  results demonstrate that our methodology provides a generalizable,
  knowledge-driven framework for scalable causal inference.
}




\maketitle

\section{Introduction}

Many computational biomedical tasks are inherently knowledge-based;
they cannot simply be ``learned'' from data but require combining
observations with structured background knowledge
\cite{alterovitz2011knowledge}. This integration between knowledge
and data becomes particularly important when addressing complex
problems such as drug safety monitoring, where rare but significant
adverse events must be detected despite their low prevalence in
general populations
\cite{pande2018causality,10.3389/fdsfr.2023.1193413}.

Knowledge graphs (KGs) are the main approach for organizing biomedical
knowledge, and can be used to represent biomedical entities (e.g., diseases, drugs,
proteins) and their relationships in a structured format
\cite{kg_def}. These graphs encode qualitative knowledge --- facts that are
either true or false --- and have been widely applied across life
sciences to represent taxonomic hierarchies, molecular interactions,
and clinical associations \cite{zhan2024application}. While KGs are very useful for
representing knowledge, they lack formal mechanisms to support
probabilistic and causal reasoning necessary for many biomedical
applications.

In clinical and epidemiological research, qualitative causal relations
can be expressed through directed acyclic graphs (DAGs)
\cite{dag_epidemiology}. These DAGs are a component of structural
causal models (SCMs) which combine causal DAGs with structural
equations \cite{Pearl_2009}. SCMs can be used to distinguish causal
effects from associations and to answer ``why'' questions, i.e., how a
probability distribution will change as a result of an intervention.
SCMs and KGs remain largely separate frameworks: although the
qualitative parts of an SCM (the DAG) can be embedded within a KG, and
KGs may encode relations that are causal or have causal implications,
it remains challenging to integrate the quantitative parts of SCMs
with KG reasoning.

Drug safety monitoring is one application that requires both knowledge
representation and causal reasoning capabilities. Post-marketing
surveillance aims to identify adverse drug reactions (ADRs) that may
not have been detected during clinical trials due to their rarity or
delayed onset \cite{trifiro2022new, Raj2019}. This surveillance depends on background
knowledge (disease and drug classifications and their
interrelationships, known disease progression, drug indications, and
known drug side effects) and observational data (frequencies of event
occurrences and co-occurrences). The surveillance task also requires
causal inference to determine whether an observed association
represents a genuine drug effect or results from confounding factors
\cite{pande2018causality,10.3389/fdsfr.2023.1193413}.

Current approaches to ADR detection include feature-based predictive
models using drug descriptors, and observational data analysis from
adverse event reporting systems and electronic health records
\cite{zhang2020adverse, fukuto2021predicting, hu2024data}. While
these methods are very useful, they do not fully utilize available
background knowledge, nor do they account for all confounding
variables.  Causal inference methods like mediation analysis
\cite{imai2010general} offer a direct methodological framework for
determining potential ADRs but rely on causal models which are often
hand-crafted and do not integrate with existing background knowledge
\cite{statins_mediation, metformin_mediation, medication_mediation}, which limits their
scalability.

We have developed a theoretical framework that integrates KGs and SCMs
which we call Causal Knowledge Graphs (CKGs). CKGs extend knowledge
graphs by incorporating probability distributions over graph nodes and
explicitly identifying relation types with causal semantics. This
integration allows us to automatically identify confounding variables
based on graph structure, can generate hypotheses that align with
domain knowledge through KG queries, and enables probabilistic
inference that respects KG semantics (in particular the hierarchical
relationships between entities).

We demonstrate the utility of our approach by applying it to ADR
detection using data from UK Biobank and MIMIC-IV. Our CKG-based
method successfully identifies known adverse drug reactions while also
discovering novel ones not previously documented. To validate these
novel findings, we apply the ADRs to the task of drug repurposing,
testing whether drugs with similar adverse event profiles share
therapeutic indications. Our results show significant improvement over
approaches using only established ADRs, confirming the value of the
ADRs we discover. The theoretical CKG framework we developed combines biomedical
knowledge representation and causal inference. it has applications
that extend beyond pharmacovigilance to any domain where both
observational data and background knowledge with causal components are
available.

\section{Methods}\label{method_section}
\subsection{Data}
\label{data}
We obtained causal relations between diseases from a Directed Acyclic
Graph (DAG) representing disease progression/sequelae ~\cite{mypaper}. The
content of this DAG was text-mined from the scientific literature, and
filtered using several methods to retain correct disease--disease
pairs. In the DAG, a causal relationship between two diseases
indicates that the causative disease can lead to the onset of the
outcome disease.

For indications, we used the high-precision subset of the MEDI-C
dataset \cite{zheng2021updated} which is based on mined data from EHR
data and multiple literature resources. In MEDI-C, drugs are mapped to
RxNorm identifiers \cite{nelson2011normalized} and diseases are
mapped to the International Classification of Diseases 9th and 10th
versions (ICD-10, ICD-9). For side effects of drugs, we utilized the
OnSIDES dataset \cite{tanaka2024onsides}, where drugs are mapped to
RxNorm and diseases are mapped to MedDRA terms
\cite{mozzicato2009meddra}. OnSIDES was text-mined from package
inserts of drugs and we used version 2.1.0 of the
dataset. Additionally, we used the OFFSIDES dataset
\cite{tatonetti2012data} as an additional evaluation set of side
effects.

We utilized two large cohorts to define the probability distributions
used for causal inference and mediation analysis: the UK Biobank, and
the MIMIC-IV dataset.

UK Biobank (UKB) represents a prospective cohort of more than half a
million participants aged 40--69 years \cite{ukb}, and reports
diagnoses of individuals using the International Classification of
Diseases (ICD). Additionally, the
UKB provides extensive data from questionnaires and verbal interviews
including data on medications taken by participants. Medications are
assigned identifiers unique to the UKB. In addition to basic
demographic and socioeconomic data, UKB provides data about smoking,
alcohol intake, and physical activity of
participants.

The Medical Information Mart for Intensive Care IV (MIMIC-IV) is a
dataset of electronic health records (EHRs) collected from 2008 to
2019 covering 364,627 individuals. It includes hospital records of
patients admitted to the Intensive Care Unit (ICU) with diagnoses
available in the International Classification of Diseases 9th, and
10th versions~\cite{icd10}. Data on prescribed medications is also
available where medications are expressed in free text form. We
obtained data on drug use from the pharmacy records. The dataset also
includes data on age, sex, ethnicity, and basic measurements like Body
Mass Index (BMI).

\subsection{Mapping of medications and conditions}
Because UKB and MIMIC-IV use different identifiers for diagnoses and
drugs, they are not directly mappable. We developed a hybrid approach
to map drugs from RxNorm to both cohorts. For each RxNorm drug, we
first attempted to find an exact lexical match between its preferred
name and drug names recorded in UKB and MIMIC-IV. If no exact match
was available, we identified as candidates any medications in UKB or
MIMIC-IV that partially matched the RxNorm name on a word-by-word
basis. To resolve these partial matches, we used the
\texttt{Llama-3-70B-Instruct} \cite{grattafiori2024llama} Large
Language Model (LLM). For each RxNorm drug, we prompted the LLM (see
Table S1) to select the most appropriate candidates from the list.

To evaluate the accuracy of the LLM-based mapping, we randomly
selected 50 drugs (25 from UKB and 25 from MIMIC-IV) for manual
review. An expert (P.N.S) examined the candidate lists for
each RxNorm drug, using the preferred name as reference. The expert
determined that a candidate was a correct match if the RxNorm drug ---
based on its preferred name --- was fully or partially represented in
the candidate, meaning the candidate correctly corresponded to the
intended drug entity, even if the names were not identical.  Based on
this manual curation, the LLM achieved a precision of $0.70$, a recall
of $0.83$, and an F1-score of $0.76$. Using this hybrid approach, we
mapped 941 drugs from RxNorm to UKB, of which 233 were based on exact
match. We also mapped 839 drugs from RxNorm to MIMIC-IV of which 541
were based on an exact match. The manually curated drug mappings and
the mapping results of the LLM are available in the Github repository.

Both OnSIDES and OFFSIDES report side effects using the Medical
Dictionary of Drug Regulatory Activities (MedDRA). On the other hand,
the causal relations between diseases in the graph we use are reported
using the International Classification of Diseases, tenth revision,
Clinical Modification (ICD-10-CM) codes. Therefore, we mapped
conditions from MedDRA to ICD-10-CM using the Unified Medical Language
System (UMLS) \cite{UMLS}. Furthermore, we used UMLS to map ICD-9
codes to ICD-10 to represent diagnoses.

\begin{figure*}[]

\centering
{\includegraphics[scale=0.3]{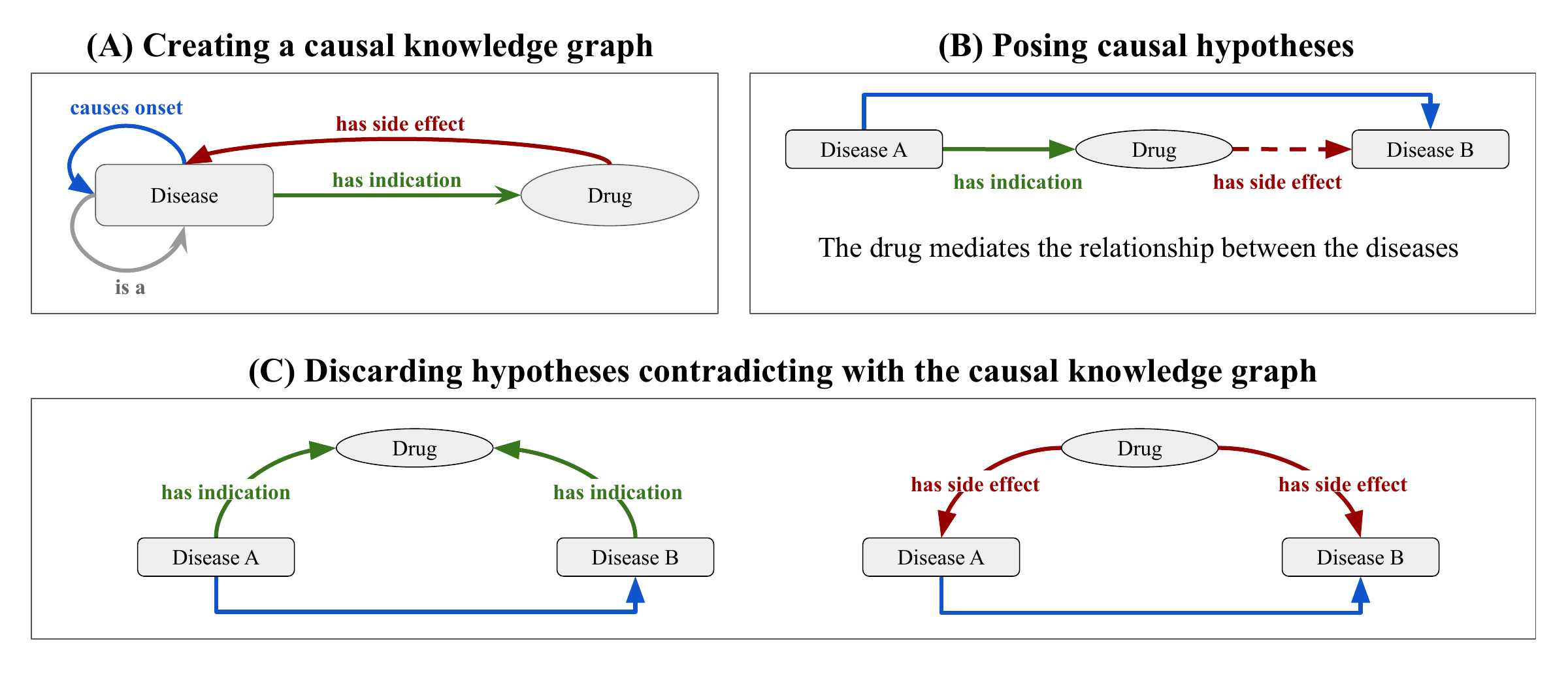}}
\caption{The process of generating hypotheses from the CKG}
\label{workflow_fig}
\end{figure*}

\subsection{Causal Knowledge Graph Construction}
\label{ckg_creation}

Knowledge can be represented in structured forms that are
interpretable by machines, such as knowledge graphs (KGs). A KG can be
represented as a tuple \[\mathcal{K} = (V, E, R)\] where $V$ are nodes
representing entities and \( E \subseteq V \times R \times V \) is a
set of directed edges labeled by relations in $R$.  KGs enable
reasoning over complex interconnections in data. In contrast,
Structural Causal Models (SCMs) \cite{Pearl_2009} offer a formal
framework specifically designed to represent and reason about causal
relationships between random variables. In SCMs, a directed acyclic
graph (DAG) is used where each node represents a variable, and
each directed edge represents a direct causal effect from one variable
to another.  SCMs further define functions for each variable $x$ as $x
= f(pa_x,u_x)$ where $pa_x$ denotes variables with outgoing edges into
$x$, and $u_x$ represents the errors due to unobserved factors.

All edges in SCMs are interpreted as causal relationships, which is in
contrast to knowledge graphs that can include multiple types of
relations. Some knowledge graphs may also include subsumption
relations that structure nodes into subsumption hierarchies \cite{pham2020constructing},
representing, for example,
diseases and their subtypes.

We introduce the concept of a \emph{Causal Knowledge Graph (CKG)},
which integrates the flexible structure of KGs with causal semantics
and a probabilistic interpretation, while adhering to the constraints
defined by KG relationships.  We formally define a Causal Knowledge
Graph as follows:

\begin{definition*}[Causal Knowledge Graph]
\label{ckg_def}
Let \( \mathcal{K} = (V, E, R) \) be a knowledge graph. Let \( \Omega
\) be a non-empty set representing the population (e.g., individuals),
and let \( 2^\Omega \) denote the power set of \( \Omega \).

A \textbf{Causal Knowledge Graph (CKG)} is a tuple:
\[
\mathcal{G} = (\mathcal{K}, R_{\text{causal}}, \Omega, f, P)
\]
where:
\begin{itemize}
    \item \( R_{\text{causal}} \subseteq R \) is the subset of
      relations that are interpreted as causal,
    \item \( f: V \rightarrow 2^{\Omega} \) is a function that assigns
      to each node a subset of the population $\Omega$,
    \item \( P: 2^{\Omega} \rightarrow [0,1] \) is a probability
      measure over subsets of \( \Omega \).
\end{itemize}
The triple \( (\Omega, 2^\Omega, P) \) forms a probability space over
the population. Each node \( v \in V \) is assigned a probability by
applying \( P \) to its corresponding subset
\( f(v) \subseteq \Omega \); that is, \( P(f(v)) \).

The function \( f \) may be subject to constraints derived from the
relations in \( R \).
\end{definition*}

In our work, we apply a constraint to ensure that \( f \) respects the
\texttt{is\_a} hierarchy between diseases by imposing the condition:
\[
\forall (u, \texttt{is\_a}, v) \in E, \quad \forall x \in \Omega,
\quad x \in f(u) \rightarrow x \in f(v)
\]

In the CKG, the probability measure $P$ associates each entity with a
random variable, enabling the interpretation of relationships between
entities as constraints on the joint distribution over these
variables. This allows us to explicitly model causal relationships
between variables and apply causal inference methods to the causal
subgraph, while preserving the semantics of selected non--causal
relations.

For this study, we created a CKG with nodes representing diseases and
drugs (the Drug--Disease Causal Knowledge Graph, DD-CKG). The nodes in
the DD-CKG are connected by the following types of relations: disease
progression, spanning 7,586 edges (extracted from a DAG of causal
disease relations); indications, comprising 20,955 edges (derived from
MEDI-C); side effects of drugs, represented by 59,119 edges (sourced
from OnSIDES); and the ICD-10 disease hierarchy, encoded as
\texttt{is\_a} relations, as illustrated in Figure
\ref{workflow_fig}A. As an example, in the ICD-10 hierarchy, ``Type 2
diabetes mellitus without complications'' (E11.9) \texttt{is\_a}
``Type 2 diabetes mellitus'' (E11). The inverse of indication edges
(interpreted as a disease leading to the prescription of a drug) and
disease progression edges together form the causal relation subset
$R_{causal}$ in the CKG. The ICD-10 hierarchy was used to impose the
\texttt{is\_a} constraint on the mapping function \( f \) as described
in Definition~\ref{ckg_def}. The population set \( \Omega \) consisted
of individuals drawn from either the UKB or MIMIC-IV.

\subsection{Generation of hypotheses}
\label{hypotheses_set}

We generated candidate hypotheses of drug-mediated disease
interactions using two sources of candidate disease progression or sequel
relations. The first source was our CKG, from which we extracted
directed pairs of the form \( x \xrightarrow{\text{causes onset}} y \)
where $x$ and $y$ are diseases. For each such pair, we queried the CKG
for drugs indicated for the source disease \( x \), forming candidate
hypotheses of the form: disease \( x \) (indication)
\(\xrightarrow{\text{causes prescription}}\) drug
\(\xrightarrow{\text{causes onset}}\) disease \( y \) (potential side
effect). As shown in Figure~\ref{workflow_fig}B, these hypotheses
represent potential cases where a drug mediates the causal
relationship between diseases. As shown in Figure~\ref{workflow_fig}B,
these hypotheses reflect potential cases where a drug mediates the
causal relationship between diseases.

The second source of hypotheses consisted of statistically
significant comorbidities between diseases identified from the
UKB cohort. We computed the relative risk (RR) between co-occurring
diseases following the approach in~\cite{hidalgo2009dynamic}. After
correcting for multiple comparisons using the Benjamini--Hochberg
procedure (\( \alpha = 0.05 \)), we retained 34{,}843 significant
associations. To orient these associations, we used diagnosis
timestamps: for each disease pair \( (x, y) \), if \( x \) was
diagnosed before \( y \) in the majority of cases in the UKB, we
interpreted the direction as \( x \rightarrow y \).

From both sources, we constructed initial hypotheses and applied
filtering to exclude to exclude hypotheses that could already be
explained by knowledge encoded in the CKG as illustrated in
Figure~\ref{workflow_fig}C.  In particular, we removed:
\begin{enumerate}
\item Hypotheses where a drug was indicated for \emph{both} diseases
  \( x \) and \( y \), which can reflect general treatment overlap
  rather than a mediating effect.
\item Hypotheses where both diseases were listed as side effects of
  the same drug, which would imply a common downstream effect rather
  than a directed causal chain.
\end{enumerate}
This process can be formulated as a SPARQL query \cite{prudhommeaux2008sparql} on the
CKG (shown in the Supplementary section 1).

As a result, we retained only hypotheses where the drug was indicated
for the source disease, could plausibly contribute to the onset of the
target disease, and there are no alternative explanations already
existing in the graph. We further excluded hypotheses that lacked
sufficient sample sizes, specifically cases where the drug or either
disease did not appear in the data, or where no individuals had all
three components of the hypothesis co-occurring (\texttt{indication =
  1, drug = 1, side effect = 1}). This filtering process yielded
12,561 hypotheses based on causal progression pairs and 81,610
hypotheses based on comorbidity pairs.

\subsection{Sample selection} \label{sample_selection}

To confirm the incidence of indications and side effects, we used the
reported codes of ICD-10 diagnoses and their dates for each individual
in the UKB (Data-Field 41270). For MIMIC-IV, we retrieved all ICD-10
and ICD-9 codes with the reported dates for each individual. In the
UKB, information about drug prescriptions includes self-reported
information where individuals were asked to provide information about
medications they were taking (Data-Field 20003). The UKB provides data
on multiple visits of individuals to their centers and prescriptions
were reported at each visit to the assessment centers. In the MIMIC-IV
dataset, we used pharmacy records of prescribed drugs and their
duration.

To obtain longitudinal data, we first excluded individuals without
follow-up data. We then excluded samples where the variables did not
follow the required temporal order: indication diagnosis, followed by
drug use, and then side effect diagnosis. Specifically, we excluded
individuals diagnosed with the outcome before either the drug or
indication. Since drugs can be taken intermittently or consecutively,
we excluded those who reported using the drug only before the
indication diagnosis.

\subsection{Confounding control}
\label{confounding_control}

To correct for pre-treatment confounders, we considered sex, age,
ethnicity, and BMI as potential confounders. Similar to other
approaches using mediation analysis in UKB~\cite{statins_mediation},
we also included education, Townsend deprivation index (TDI), alcohol
use, smoking, and physical activity when analyzing data in UKB; this
information is not available in MIMIC-IV. To account for disease
severity which in not directly observed, we used the number of
comorbid diseases and the number of prescribed drugs as proxies
because they can correlate with disease
severity~\cite{forslund2021patterns}.

\begin{figure}[htbp]
    \centering
    \includegraphics[scale=0.4]{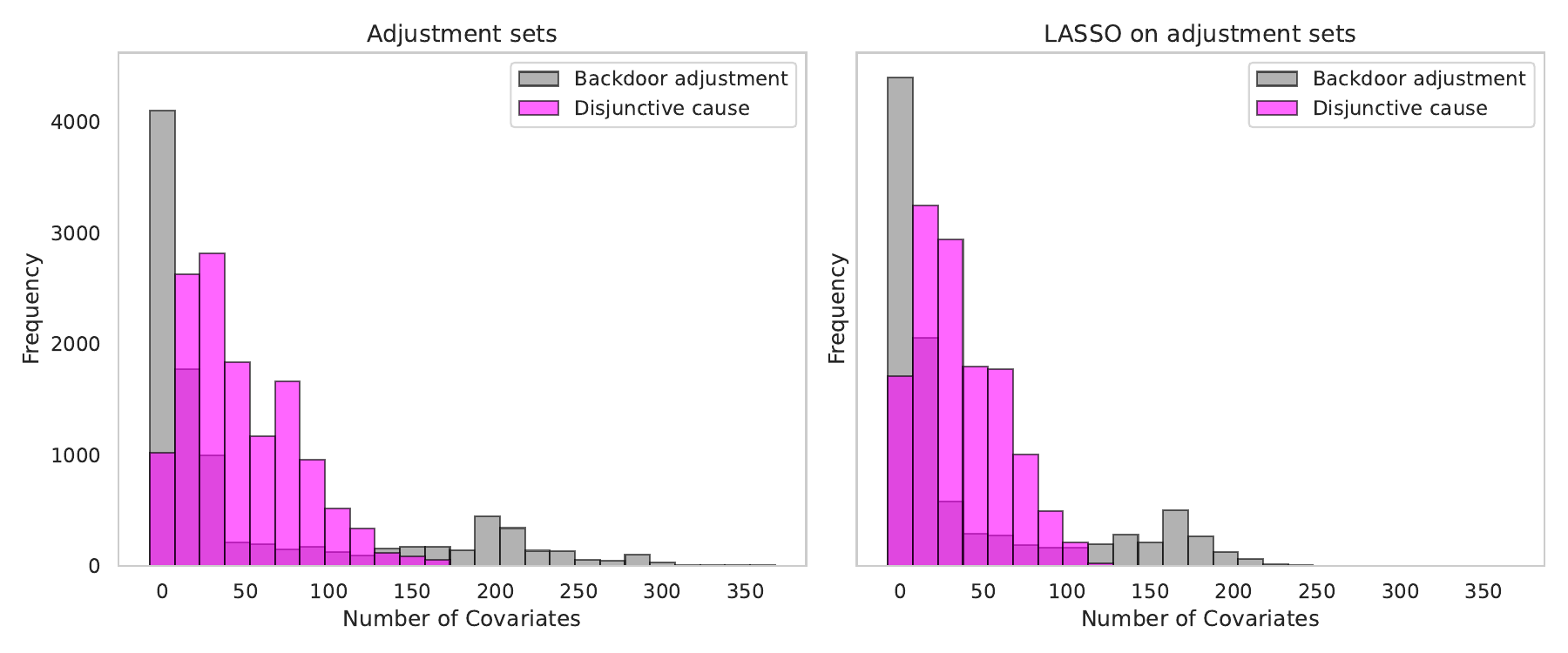}
    \caption{Distribution of the number of covariates per method.}
    \label{covariates_dist}
\end{figure}

We control for concomitant drug use and comorbid conditions using the
information in our CKG. Specifically, we capture comorbid conditions
by disease--disease causal edges, and concomitant drug exposures via
indication edges. Based on the edges with relations belonging to
$R_{causal}$ in our CKG, we applied two criteria to identify
adjustment sets: (1) the backdoor adjustment criterion which
identifies confounders through backdoor paths (i.e., paths from the
treatment to the outcome that go through confounders) in a causal
graph \cite{pearl2018book}; and (2) the disjunctive cause criterion
which selects causes of the treatment, or the outcome, or both
\cite{vander2011new}.  For each hypothesis independently, we first
added a causal edge from the drug to the potential side effect then
identify confounders through applying one of the criteria. In the
adjustment sets, it sometimes happened that both a disease and its
more specific child (according to the ICD-10 hierarchy) appeared
together. To avoid redundant adjustment, we pruned these by keeping
only the parent disease and removing the child disease.

We used the DAGitty package in R \cite{textor2016robust} to apply the
backdoor adjustment. 
we only considered the first 1,000 adjustment sets returned by
DAGitty. Among the returned sets, we selected the one with the lowest
cardinality. If multiple sets shared the minimum cardinality, we
retained the first encountered.

The number of selected covariates can be large
(Figure~\ref{covariates_dist}). We used the Least Absolute Shrinkage
and Selection Operator (LASSO) to select covariates from the
identified adjustment sets.
The estimated coefficients $\hat{\beta}$ minimize the following
optimization objective:
\[
\frac{1}{2n} \sum_{i=1}^{n} \left( y_i - X_i^\top \beta \right)^2 +
\lambda \sum_{j=1}^{p} |\beta_j|,
\]

where $y_i$ is the outcome (side effect), $X_i$ is the vector of
predictors (covarites in an adjustment set) for observation $i$,
$\lambda \geq 0$ is a regularization parameter,
and $p$ is the number of covariates in the adjustment set.
Covariates with non-zero coefficients in $\hat{\beta}$ are considered
selected. This leads to smaller, more interpretable sets of covariates
\cite{ye2021variable, urminsky2016using}.
Following the methodology in~\cite{urminsky2016using}, we fitted two
separate LASSO models --- one for the indication and one for the side
effect --- and used the union of the variables selected in either
model for adjustment. To optimize the models, we selected the
regularization parameter ($\lambda$) that minimized the 10-fold
cross-validation error. As depicted in Figure~\ref{covariates_dist},
LASSO shrinks the number of selected covariates.

\subsection{Statistical analysis}\label{stat_analysis}

To study the mediating effects of drugs between indications and
outcomes, we use causal mediation analysis grounded in the potential
outcomes framework. Let \( T \) denote the treatment (indication),
\( M \) the mediator (drug), and \( Y \) the outcome (side effect). We
are interested in the {\em natural indirect effect (NIE)}, which
quantifies the part of the effect of \( T \) on \( Y \) that operates
through the mediator \( M \) rather than through direct paths.

Following the definition in \cite{pearl2014interpretation}, the NIE
is given by:
\[
  \text{NIE} = \mathbb{E}[Y_{0, M_1} - Y_{0, M_0}],
\]
where \( Y_{t, M_{t'}} \) denotes the potential outcome if treatment
were set to \( t \) and the mediator to the value it would have had
under treatment \( t' \).  Potential outcomes represent the values the
outcome would attain under specific interventions.  This captures the
change in the outcome caused by shifting the mediator from its
untreated to treated value, while keeping the treatment fixed at the
baseline \( T = 0 \).

To estimate the NIE, we use the R package \texttt{mediation}
\cite{mediation_r} which estimates the potential outcomes using
regression models. In the \texttt{mediation} package, the NIE is
identified as the Average Causal Mediation Effect (ACME). The ACME is
identified under the sequential ignorability assumption, namely that,
conditional on the observed pre-treatment covariates \( X \), (i) the
treatment \( T \) is independent of all potential mediator and outcome
values, and (ii) the observed mediator \( M \) is independent of all
potential outcomes given \( T \) and \( X \). We adjusted for
pre-treatment confounders by including them as covariates in the
regression models and included interaction terms ($T \times M$) when
statistically significant (\( p < 0.05 \)). Estimation was performed
with 1,000 quasi-Bayesian simulations with
heteroskedasticity-consistent standard errors. Finally, we applied
multiple testing correction via the Benjamini--Hochberg procedure on
the mediation results with \( \alpha = 0.05 \).

\subsection{Prediction of shared indications}
\label{indication_similarity}

As intially proposed in \cite{campillos2008drug}, side effect
similarity can be used to predict if two drugs share
indications. Following the approach in \cite{tatonetti2012data,
  tanaka2024onsides}, we repeated the analysis by computing the
pairwise Tanimoto coefficient score for drugs based on their side
effects. That is, for two drugs $A$ and $B$ and their corresponding
sets of side effects $SE_{A}$ and $SE_{B}$, their similarity
$Sim(A,B)$ is calculated by:
\[
Sim(A, B) = \frac{|SE_A \cap SE_B|}{|SE_A \cup SE_B|},
\]
We compared different sources using side effects either reported by
OnSIDES, obtained from our analysis, or the union of both. For each
configuration, we computed pairwise drug similarity scores using the
Tanimoto coefficient and applied $z$-score normalization to the
resulting similarity matrix. To evaluate whether these scores could
predict whether two drugs share a common indication (a binary
outcome), we assessed the predictive performance using the area under
the receiver operating characteristic curve (ROC AUC).

\subsection{Side effect evaluation}
\label{evaluation_methods}

We used the OnSIDES and OFFSIDES datasets to construct a reference set
of known drug-–side effect pairs.  For each drug--side effect pair
identified by our mediation analysis, we classified it as a true
positive if it appeared in the reference set, and as a false positive
if it did not.  False negatives were defined as drug--side effect
hypotheses that (i) were generated in Section~\ref{hypotheses_set},
(ii) appear in the reference set, but (iii) did not reach statistical
significance in our analysis.

\section{Results}

\subsection{A framework for knowledge-based causal mediation analysis}
We developed a method to identify post-marketing adverse effects of
drugs from large-scale observational cohorts while controlling for
confounding. We model this task as a mediation problem: testing
whether a drug mediates the effect between an indication and a side
effect. 
We then apply causal mediation analysis to identify drugs that
significantly mediate associations between diseases.

To generate plausible hypotheses and control for confounding at scale,
we first constructed a Causal Knowledge Graph (CKG)
(Section~\ref{ckg_creation}).  We define a CKG as a structure
consisting of a knowledge graph, a subset of relations marked as
``causal'', a probability space, a mapping between knowledge graph
nodes and events in the probability space, and a set of constraints
that ensure that the relational semantics in the knowledge graph is
reflected in the probability space. We build a Drug--Disease Causal
Knowledge Graph (DD-CKG) that enables us to identify mediating effects
of drugs. The DD-CKG integrates disease progression, drug indications,
side effects of drugs, and hierarchical relations from the ICD-10
hierarchy. We consider edges representing disease progression and the
inverse of indication edges (i.e., that a disease diagnoses may lead
to the prescription of the drug) as causal. We empirically assign a
probability distribution to the CKG based on longitudinal cohort data
in UK Biobank (Section~\ref{data}).

The DD-CKG allows us to automate three key steps in mediation
analysis: (i) posing hypotheses consistent with background knowledge,
(ii) identifying confounding structures, and (iii) constraining the
probability distribution to respect the KG's prior semantics. We focus
on pairs of diseases \( D_1 \) and \( D_2 \) that satisfy the
following conditions: (a) a drug \( M \) is prescribed for \( D_1 \);
(b) \( M \) is not indicated for both \( D_1 \) and \( D_2 \); and (c)
\( M \) does not list both \( D_1 \) and \( D_2 \) as side effects
(Figure~\ref{workflow_fig}C).  Because there are many disease pairs,
we focus on two sets of disease pairs that may indicate disease
progression: (1) the disease pairs that are explicitly linked in
DD-CKG with a disease progression edge ({\tt causal set}, 12,561
disease--disease pairs), and (2) diseases that are significantly
co-morbid in UK Biobank and where one disease more often occurs before
the other disease ({\tt comorbidity set}, 81,610 disease--disease
pairs).

\subsection{Mediation analysis reveals highly concordant side effects}

To apply causal mediation analysis, we used two observational cohorts,
UK Biobank (UKB) and MIMIM-IV (Section~\ref{sample_selection}). We
used these cohorts to assign a probability distribution to the DD-CKG,
and applied logical constraints from the DD-CKG to this distribution to
make it consistent with the semantics of subsumption relations
(Section~\ref{ckg_creation}).
To adjust for confounding, we considered both demographic and
socioeconomic factors, as well as two additional adjustment criteria:
the backdoor and disjunctive cause criteria, applied to the DD-CKG (see
Section~\ref{confounding_control}). For each hypothesis, we computed
the Average Causal Mediation Effect (ACME) \cite{mediation_r} of the
drug, adjusting for the identified confounders. This analysis allows
us to assess whether the drug significantly mediates the relationship
between indications and possible side effect, indicating a direct
effect of the drug on the outcome. 

\begin{table*}[ht]
\centering
\caption{Summary of mediation analysis results by hypothesis set and adjustment method. \textbf{N/A} indicates hypotheses that could not be tested; \textbf{Insig} denotes non-significant p-values; \textbf{+/- ACME} represents significant Average Causal Mediation Effects (ACMEs). We used Benjamini--Hochberg correction for all tests.}
\resizebox{\textwidth}{!}{%
\begin{tabular}{|l|l|r|r|r|r|r|r|r|}
\hline
\textbf{Hypotheses set} & \textbf{Adjustment method} & \textbf{N/A} & \textbf{Insig.} & \textbf{+ ACME} & \textbf{- ACME} & \textbf{Precision} & \textbf{Recall} & \textbf{F1} \\
\hline
Causal & Backdoor adjustment & 8,459 & 2,433 & 1,200 & 1,578 & 0.876 & 0.084 & 0.153 \\
Causal & LASSO of backdoor adjustment & 3,226 & 4,908 & 3,237 & 2,299 & 0.905 & 0.233 & 0.371 \\
Causal & Disjunctive cause & 10,694 & 1,269 & 860 & 847 & 0.859 & 0.059 & 0.110 \\
Causal & LASSO of disjunctive cause & 2,128 & 5,287 & 3,910 & 2,345 & 0.907 & 0.282 & 0.431 \\
\hline
Comorbidity & LASSO of disjunctive cause & 34,643 & 20,802 & 30,724 & 13,158 & 0.749 & 0.282 & 0.410 \\
\hline
\end{tabular}
}
\label{tab:merged_summary}
\end{table*}


We computed the ACME of both sets of disease -- disease pairs (the {\tt
  comorbidity set} and the {\tt causal set}), using different methods
to adjust for confounding. We find that selecting confounders using
the disjunctive cause criterion followed by selection through LASSO
resulted in most testable hypotheses on the {\tt causal set}, and due
to the larger size of the {\tt comorbidity set}, we only applied this
confounder control on the {\tt comorbidity set}.  A positive ACME
indicates that the drug may contribute to the occurrence of the
outcome disease, while a negative ACME indicates that the drug may
reduce the likelihood of the outcome disease. We focus only on
positive ACME and compared resulting drug--disease pairs to existing
adverse event databases. Table~\ref{tab:merged_summary} shows the
results of the ACME and the comparison to existing databases.

We find that the mediation analysis over the DD-CKG can reveal both known
and novel drug effects. Using edges that are included in the DD-CKG as
candidates, we find that most identified effects are already known and
contained in a drug effect database (precision up to $0.907$). This is
expected as the pairs included in the DD-CKG are supported by literature
and therefore likely correspond to established effects. Using
significantly comorbid diseases in UKB, on the other hand, revealed
more candidate drug effects (30,724 significant positive effects), the
same recall, but lower precision (i.e., more potentially novel
effects).

To further evaluate the drug effects we identify from the
\texttt{comorbidity set}, we used them to compute side-effect
similarity between drugs and tested whether drugs with higher side
effect similarity share indications.
While using only drug effects predicted by the causal mediation
analysis yields a lower performance (ROCAUC: $0.604$) than using drug
effects from the OnSIDES database (ROCAUC: $0.620$), combining OnSIDES
and our predicted drug effects significantly improves the prediction
of shared indications(ROCAUC: $0.632$, $p \ll 0.001$, Mann Whitney U
test).

\begin{figure*}
  \centering
  \includegraphics[scale=0.32]{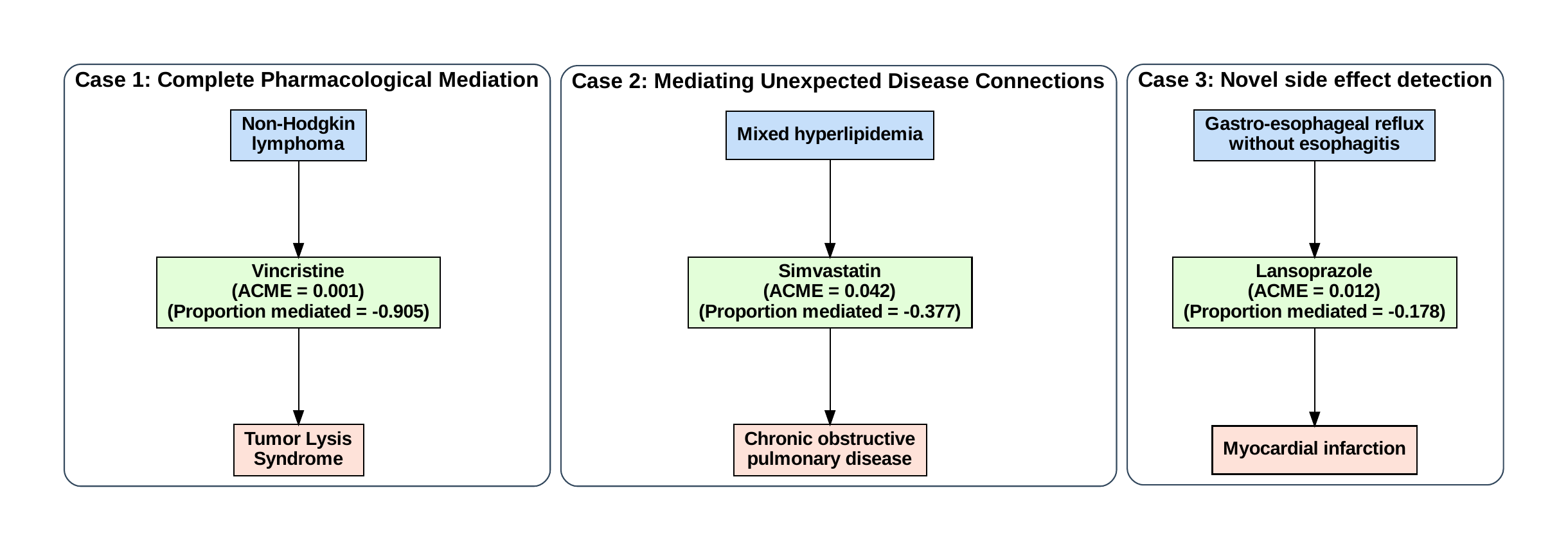}
  \caption{Mediation cases representing different explanations of
    inter--disease relationships.} 
  \label{result_fig}
\end{figure*}

\subsection{Ethical approval}
Research using UK Biobank data was approved by the Institutional
Bioethics Committee (IBEC) at KAUST under approval number 22IBEC054.

\section{Discussion}
\subsection{Mediation analysis explain disease--disease relationships}

Using the results of the mediation analysis on the \texttt{comorbidity
  set}, we can find explanations for disease--disease
relationships. We discuss two illustrative cases below.  As shown in
Figure \ref{result_fig}, we observed that the drug
\textit{Vincristine} fully mediates the causal effect of
\textit{Non-Hodgkin lymphoma} on \textit{Tumor Lysis Syndrome} where
the direct effect was insignificant. The estimated average causal
mediation effect (ACME) was $0.001$, with a proportion mediated of
$-0.905$, indicating that nearly the entire effect is derived through
the drug. The negative sign of the proportion is due to differing
signs in the direct and indirect effects, a phenomenon known as a
suppression effect \cite{mackinnon2000equivalence}. This result
reinforces the clinical understanding that Tumor Lysis Syndrome arises
primarily as a complication of chemotherapy agents such as Vincristine
\cite{williams2019tumor}.

As shown in case 2 in Figure \ref{result_fig}, we found
\textit{Simvastatin}, a cholesterol-lowering drug, to mediate the
causal link between \textit{Mixed hyperlipidemia} and \textit{Chronic
  obstructive pulmonary disease (COPD)}. The ACME was $0.04177$ and
the proportion mediated was $-0.377$. Although rare, statins can cause
interstitial lung disease; however, COPD itself has not been
reported~\cite{huang2013statin, Fernndez2008}. The most common causes of COPD are primary and secondary smoking \cite{LaniadoLaborn2009}.  We already control for this as a confounder in the UKB data, supporting the hypothesis that the occurrence of COPD may indeed reflect mediation through simvastatin. This case shows how a
seemingly non-intuitive causal link between metabolic and respiratory
conditions can be elucidated through the mediating influence of
pharmacological exposure.

Where the outcome disease was not a recognised ADR, nor a common
progression of the initial disease, we established from expert
knowledge if this could be a plausible side effect of the drug given
its physiological mode of action and metabolism. We then searched the
literature manually for publications reporting such ADRs. Using this
expert curation we find some unexpected or very rare associations, for
example between gastroesophageal reflux disease (GERD) and myocardial
infarction (Case 3 in Figure~\ref{result_fig}). These effects are
mediated in the former case by lansoprazole and omeprazole, both
widely used proton pump inhibitors \cite{Strand2017}. Similar
unexpected associations between nephropathy and myocardial infarction
were found to be mediated through ACE inhibitors lisinopril,
enalapril, and perindopril and ramipril \cite{IzzoJr2011}, suggesting
a systematic drug class side effect. ACEs are in widespread use for
both early and advanced diabetic and non-diabetic nephropathy
\cite{Bhandari2022}; instances of myocardial infarction are reported
in several studies, including significant association with worsening
heart failure \cite{NaTakuathung2022}.  The increased risk of
myocardial infarction associated with protein pump inhibitors is not
included in drug labels \cite{Elias2019} and has not been implicated
in a recent analysis of ambulatory healthcare data
\cite{Ma2020}. However, interestingly, the adverse effect was
previously identified in a large EHR data mining study of 1.8 million
individuals with 20 million patient interactions in the Stanford
University Hospital and Clinics system \cite{Ariel2019, Shah2015}.

\subsection{Disease severity as a confounder}

We used the number of comorbid conditions and the number of prescribed
medications as proxy variables for disease severity (which is not
directly observable in our dataset). Nevertheless, we noted that in
the curated set of potentially novel adverse drug events, the outcome
disease was a frequent progression of the initial disease; similarity,
the outcome disease would sometimes indicate an increased severity of
the initial disease.  In the 900 false positives we evaluated
manually, we find that 6\% plausibly represented disease progression
or increased severity rather than mediation through the drug.  Here,
the unobserved confounding occurs when a drug $M$ is prescribed for a
condition which is not mitigated by the treatment, but which then goes
on to progress with unusual severity.  In other, less severe cases
where the treatment is effective, there will be no outcome disease
associated with the treatment. This will result in $M$ appearing to
mediate the appearance of $D_2$ because it is only observed when $D_2$
is also present.

One example is \emph{Disorder of kidney and ureter} (N28.9) as an
initial disease which is associated with \emph{Malignant renal
  neoplasm} (C64) for eleven drugs. All drugs are used in the
treatment of hypertension, and none are thought to cause
malignancy. However it is well established that high blood pressure,
long term dialysis, renal failure, and other risk factors are
associated with malignancy \cite{Chow2010}, and we suggest that this
pattern of inferred mediation is due to the severity of these cases.
We also see well known progressions such as \emph{Mixed
  hyperlipidemias, Hyperlipidemia, Hyperglycemia} as well as
\emph{Alcohol dependence} all resulting in \emph{Fatty change of the
  liver} \cite{Israelsen2024}, but flagged as being drug mediated.

\subsection{Causal Knowledge Graphs}

A major contribution of our work is the development of a novel class
of Causal Knowledge Graphs (CKGs), which extend traditional knowledge
graphs with formal semantics that enable causal inference while
preserving relational semantics, supports deconfounding via explicitly
marked causal edges, and facilitates the formulation of causal
hypotheses directly aligned with background knowledge encoded in the
graph.  There is prior work on combining knowledge graphs with causal
inference. One set of methods considers knowledge graphs where
relations have a causal interpretations, i.e., where the graphical
model coincides entirely or in parts with the knowledge graph
\cite{info14070367,tan-etal-2024-enhancing-fact}; these methods focus
on inference of causal relations using the relational semantics of a
knowledge graph but do not integrate the graph with the probability
distribution associated with a causal model.  CauseKG \cite{causekg}
is an approach to integrate the probability distribution of a causal
model with relational semantics, but is limited to the notion of
``identity'' between graph nodes and considering sub-properties.

We establish a link between the knowledge graph semantics and a
probability space, which allows for a deeper integration between
knowledge graph semantics and the probability distribution associated
with the causal model. It is this integration that allows us to
exploit subsumption relations in the knowledge graph (between
diseases) to constrain the probability distribution. Furthermore, we
do not identify the knowledge graph with the graph in the causal
model, but focus on only a subset of relations in the knowledge
graph. This allows us to use other relations, and rules, to
deductively infer edges in the graph, and then generate the causal
graph from the asserted and inferred edges. In the future, this
approach of combining formal semantics of knowledge representation
languages and causal semantics may be further extended to more
expressive knowledge representation frameworks.



\subsection{Application to longitudinal cohorts and the Electronic
  Health Record}

The causal mediation framework we propose is general and can be
applied to other longitudinal cohorts.  The application of our
framework requires longitudinal data including diagnoses and drug
prescriptions, and the ability to map the diagnoses and prescriptions
to standardized identifiers that are included in our knowledge
graph. This standardization is a limiting factor especially in
biobank-style cohorts where information from multiple different
sources (e.g., primary care, hospital records, and self-reported
information from surveys) is integrated.
We standardized medications using a combination of a lexical approach
and a Large Language Model (LLM), which resulted in several incorrect
mappings between drugs reported in UK Biobank and our CKG. Our mapping
approach can potentially be extended in the future by considering
other LLMs or other approach to map labels to a structured vocabulary.

In our work, drug incidence was based on self-reported medication use
in UKB, which lacks accurate information on prescription timing and
dosage. More accurate information could be obtained from the
Electronic Health Record (EHR). 
Despite these limitations, we found that many of the mediation effects
detected were consistent with known adverse events reported in
resources such as OnSIDES and OFFSIDES. 

Furthermore, in our CKG, we used ICD-coded diagnoses to represent
observed phenotypes. While these ensure clinician validation,
transient or mild side effects are not observed. To include this
information, electronic health records could be mined for milder
effects using text mining approaches \cite{Slater2021, Iyer2014}, and
using vocabularies that can capture these effects.

\section{Conclusion}

We have developed a novel approach to identify rare adverse drug events
from longitudinal health data. Our approach is enabled by a novel
framework that combines knowledge graphs with causal models, which we
call Causal Knowledge Graphs (CKGs). CKGs enable the identification
and control of confounding variables, allow entailment of causal
relations using deductive inference, and can constrain a probability
distribution with background domain knowledge. These properties of
CKGs together with the availability of large amounts of biomedical
domain knowledge in the form of biomedical knowledge graphs, as well as
large longitudinal cohorts, allows us to find rare adverse events with
low effect size that have been missed in other studies. Moreover, our
analysis relies on standard identifiers used in health records and
standard representation formats for biomedical knowledge; it therefore
has the potential to be extended to other applications of causal
inference where structured domain knowledge and observational data are
both available, and the observations can be linked to entities in the
domain knowledge.

\section*{Acknowledgements}
We acknowledge the use of computational resources from the 
KAUST Supercomputing Core Laboratory.

This research has been conducted using the UK Biobank Resource under
Application Number 31224.

\section*{Funding}
This work was supported by funding from King Abdullah University of
Science and Technology (KAUST) Office of Sponsored Research (OSR)
under award numbers URF/1/5041-01-01, FCC/1/5932-10-02,
FCC/1/5932-11-01, FCC/1/5932-12-10, REI/1/4938-01-01,
REI/1/5659-01-01, and REP/1/6338-01-01.

We acknowledge funding from King Abdullah University of Science and
Technology (KAUST) -- KAUST Center of Excellence for Smart Health
(KCSH), under award number 5932, and by funding from King Abdullah
University of Science and Technology (KAUST) -- Center of Excellence
for Generative AI, under award number 5940.

\section*{Availability of data and software}
\sloppy
\url{https://github.com/bio-ontology-research-group/Mediation-Analysis-using-Causal-Knowledge-Graph}.

\bibliography{sn-article}
\end{document}